# Predicting Wind Turbine Power Using Machine Learning and Weather Forecasts

Khivishta Boodhoo[a], Isaac Triguero[b c d], Josh Plumbly[e], Bruce Nicolson[e], Nicholas Watson[f*]

[a] Low Carbon Energy and Resources Technologies Research Group, Faculty of Engineering, University of Nottingham.
[b]Computational Optimisation and Learning Lab, School of Computer Science, University of Nottingham, Nottingham, United Kingdom
[c]Department of Computer science and Artificial Intelligence,University of Granada, Spain
[d]DaSCI Andalusian Institute in Data Science and Computational Intelligence, Granada, Spain
[e]Intelligent Plant, Aberdeen, Scotland
[f]School of Food Science and Nutrition, University of Leeds, Leeds, United Kingdom.
[*]Corresponding Author

## Abstract

Offshore wind turbines are widely used to generate renewable energy, but their maintenance can result in decreased efficiency due to forced shutdowns. Accurate wind turbine power predictions can identify periods of low power that would be ideal for scheduling maintenance. However, the effects of data volume, feature selection, and data preprocessing on the performance of such power prediction models have not been thoroughly studied. Besides, current models have limited transferability between different wind turbines. Therefore, this study developed a baseline Linear Regression for performance comparison with a more complex Artificial Neural Network model to predict the power output of a wind turbine, using weather conditions only to enhance applicability. A range of data preprocessing techniques were studied, and models were trained on one month and one year of data to determine the effects of data preprocessing and volume on model performance. Feature selection was explored using a Random Forest Regressor. The best results from the different models showed that the Artificial Neural Network models provided the highest accuracy, with an $R^2$ score of 0.98 and a low Mean Absolute Error of 194, when compared with the baseline model ($R^2$ score of 0.94 and Mean Absolute Error of 441). The model performance is comparable to the range of results in past studies, with the advantage that the proposed method leverages a separate weather dataset from a nearby weather station, enabling future applications for similar wind turbines in different locations. The Artificial Neural Network model was then used to identify 4-h periods of low power predictions over 2 months (simulating application for future periods), providing power output savings of approximately 2000 kW for each maintenance event.

**Keywords:**

artificial neural network; machine learning; wind turbine power prediction; energy production optimisation.

# 1 Introduction

With the ever-growing demand for energy, renewables play a significant role in decarbonising the world's economy and slowing climate change. Wind energy is one of the most prominent renewable energy sources with rapid expansion since 2000 (International Energy Agency, 2021)(De Alencar et al., 2017). In the United Kingdom (UK), wind energy accounted for 24% of total electricity generation (including renewables and non-renewables) in 2020 (Department for Business Energy & Industrial Strategy, 2021), growing in the first quarter of 2023 (Watson, 2023) to 39%. Offshore Wind Turbines (WTs) account for more than 13% of this wind energy (International Trade Administration (U.S), 2022). Offshore WTs are continuously being built because of advantages such as stronger and more consistent wind conditions at sea, fewer visual disturbances, fewer noise impacts on the population, and easier transportation and deployment of larger units (Stetco et al., 2019).

However, the rapid expansion of WTs in terms of the number of installations, turbine size, and rated capacity has a huge impact on their operational and maintenance costs (Verma, 2012), especially for offshore WTs. For them, operation and maintenance activities typically account for a significant proportion of the overall costs (e.g., 25–30% of the total lifecycle costs) (Powers, 2013), since they are both technically and logistically complex. When maintenance is performed, the WT must be shut down for technical and safety reasons, often for a long period; the typical routine maintenance time is approximately 40 hours (h) per year (European Wind Energy Association, 2009). Moreover, for offshore WTs, the startup and shutdown procedures might not be reliably performed under all conditions (Pape and Kazerani, 2020). Offshore WTs also require more inspections and repairs than onshore ones and take a longer time to access (Bocewicz *et al.*, 2024). For these reasons, it can result in the power output of the WT being disrupted and the overall output efficiency of the WT decreasing. To avoid this, scheduling routine or preventive maintenance when the WT produces little to no power is desirable, as this will not only minimise the impact on turbine power output but can also optimise the time of the operators performing the maintenance.

In this context, accurate and precise WT power predictions are useful for making efficient administrative and operational decisions in the wind energy sector (Ahmadi and Khashei, 2021).

For instance, power predictions for several days in the future allow maintenance engineers to detect periods of time (e.g., 1 to 6-h windows) of low power output suitable to deploy operators to perform maintenance on a WT. Simultaneously, they can be alerted about any problems that the WTs may have if there is a huge disparity between the predicted power output and the actual power output. In addition, with the recent increase in the use of renewable energies like WTs (which are inherently variable in power output), it is necessary to have accurate power predictions when connected to the grid, as any unanticipated change can cause an imbalance between supply and demand. Accurate power forecasting, however, can be challenging due to the uncertainty and volatility of wind speed in WT performance (Zhang *et al.*, 2019) (Yürek, Birant and Yürek, 2021). In addition, with the current rapid development and growth in wind generation, there is a necessity for power forecasting, which must be more accurate and reliable (Demolli *et al.*, 2019). More recently, Dhungana (2025) explicitly linked ML-based wind-power forecasting with maintenance planning by comparing nine model types using temporal, meteorological and curtailment variables.

To investigate the power predictions for WTs, it is important to understand that the amount of power produced greatly depends on the meteorological conditions at the site of a WT (Anisha et al., 2021). For this purpose, data-driven modelling approaches are well suited over physical/mechanistic modelling methods (Morshedizadeh *et al.*, 2018). Data-driven approaches analyse historical data of the specific system under investigation and can identify characteristics of the system and make future predictions, whereas physical approaches describe the behaviour of the system using equations based on a theoretical understanding of underlying mechanisms (Kite, 2021). Since WTs are multifaceted systems, all the intricate underlying principles and operation of WTs need not be known. In addition, they are normally equipped with a comprehensive Supervisory Control and Data Acquisition (SCADA) system (collecting a large volume of data), making data-driven modelling approaches most suitable.

A SCADA system is used on WTs to collect data and send them to a central computer for monitoring and control (Wang, Sharma and Zhang, 2014). However, the data points collected by a SCADA system can sometimes be erroneous or abnormal. Before using any kind of modelling approach, preprocessing the raw data is essential (Goh *et al.*, 2021). This process includes removing any noise and addressing data inconsistencies and missing values (Singh et al., 2021). SCADA systems also collect a multitude of variables (also called features or parameters), some of which may be irrelevant or redundant to building an effective and efficient predictive model (Morshedizadeh *et al.*, 2018). Another key consideration is that multiple streams of data may not be synchronised in the same time domain, which often limits their usability (Soraghan, 2020).

Tao et al., (2019) developed a grey correlation algorithm to extract relevant features from a massive and high-dimensional dataset collected by a WT SCADA system so that any overfitting phenomenon was reduced during subsequent modelling. Qiao et al., (2021) emphasised the need to preprocess SCADA data for WTs before making any predictions and used dispersion analysis and a binning algorithm on a power curve to remove abnormal data. Zhao et al., (2012) used a Kalman filter on their model inputs, which were Numerical Weather Predictions (NWP); weather forecasts made using a set of fluid flow equations, before inputting these into a data-driven model known as an Artificial Neural Network (ANN) to forecast wind power. The use of the Kalman filter improved the performance of the model. Manobel et al., (2018) applied a Gaussian process for detecting and removing outliers from SCADA data, which improved the performance of their ANN model for predicting the power curve of a WT. Lin and Liu., (2020) used Pearson product-moment correlation coefficients to incorporate only relevant input features and an Isolation Forest to remove outliers in their work on wind power forecasting. Khan et al., (2019) applied a Principal Component Analysis (PCA) method to input datasets obtained from the National Renewable Energy Laboratory (NREL) in the US before developing wind power forecasting models. Liu et al., (2023) used a digital filter to reduce noise before performing modelling on their WT data. To summarise the preprocessing needs for WT datasets, Hanifi et al. (2020) reviewed the literature on different WT forecasting model work and concluded that data preprocessing challenges still exist when developing Data Driven Models (DDMs) because the datasets vary drastically in data volume and feature type. Currently, no standard data preprocessing methodology exists which is suitable for all WTs. Kirchner-Bossi, Kathari and Porté-Agel (2024) similarly combined weather-model outputs with data-driven forecasting, demonstrating the value of a broad meteorological predictor space for intra-day and day-ahead horizons.

For any DDM, once the data has been preprocessed, it is necessary to determine the most suitable modelling methodology. Machine Learning (ML), a core component of data driven modelling is the most commonly used methodology. Current ML methods comprise both traditional statistical techniques and more complex ones (Foley *et al.*, 2012) (Mandzhieva and Subhankulova, 2022). Some examples of statistical methods include Linear Regression (LR), Autoregressive Moving Average (ARMA) and Autoregressive Integrated Moving Average (ARIMA) models (Torres and Alfonso, 2008). However, according to Demolli et al., (2019), classical statistical methods, are not the preferred methods for wind power forecasting, as they cannot adapt themselves to nonlinear wind data and cannot process large amounts of data easily. In addition, there is also a challenge with prediction error increasing in statistical methods as forecasting time increases (De Alencar et al., 2017).

More complex ML methods can identify patterns in multidimensional data and provide robust models from large datasets (sample and feature number) with some tolerance to outliers and noisy data (Montáns *et al.*, 2019). Different studies have been conducted to predict the power output for WTs using ML. Yurek, Birant and Yurek, (2021) proposed a two-staged technique including a clustering method (K-Means) followed by regression and tree-based algorithms to predict WT power. Singh et al., (2021) demonstrated the capability of a Gradient Boosting Regressing-based Ensemble algorithm and found it to have better accuracy in comparison to Random Forest, K-Nearest Neighbours, Decision Tree, and Extra Tree Regression algorithms for predicting power output. Malakouti, (2023) found that an Extra Tree ML algorithm worked best to predict power output. Ensemble models such as boosting, bagging, RF, and XGBoost were also investigated by Alkesaiberi, Harrou and Sun, (2022) to predict power output. Support Vector Machines (SVMs), another type of ML, have often been used in other WT application areas, including fault detection and condition monitoring (Powers, 2013). Santos et al., (2015) proposed an SVM classification-based method to identify several types of faults related to rotor blade imbalance and misalignment (Stetco et al., 2019). Although these studies are encouraging, questions remain regarding the most suitable ML algorithm to use and how results from one WT dataset can be used to study others.

In recent years, a subcategory of ML known as deep learning has experienced increased use in the wind energy industry. Deep learning uses a multi-layered structure, nonlinear transformations, and can model abstractions at a high level in large datasets (Dargan *et al.*, 2020). This can provide better model performance in terms of feature extraction and model generalisation than other ‘shallow’ machine learning methods (Chen *et al.*, 2021). ANNs with multiple layers form part of deep learning, although the minimum number of hidden layers that an ANN should have to be considered as a deep neural network is still ambiguous. Nevertheless, ANNs with multiple layers have been used in WTs for wind speed and power forecasting, design optimisation, fault detection and control (Marugán *et al.*, 2018). Delgado and Fahim, (2021) developed a Long Short-Term Memory (LSTM - recurrent neural network for sequential forecasting) prediction model to provide a short-term prediction of wind and power generation. López et al., (2018) proposed using LSTM blocks as hidden layer units in echo state networks to predict wind power, using historical wind power and NWP data. Meka et al., (2021) proposed a temporal convolutional network, a method that uses casual convolutions and dilations so that it is adaptive for sequential data, to provide an accurate multistep forecast of total wind power. Peiris et al., (2021) used an ANN based on the Levenberg–Marquardt algorithm (second-order gradient-based technique model training) to predict future wind power generation from projected climate scenarios. Ahilan, Sujesh and Yarrapragada, (2023) used a deep neural network (optimised with a hybrid Bird Swarm Merged Seagull (BSMS) algorithm) to predict WT power. Supplementary Table 1 provides an overview of previous studies using ML models to predict WT power output. Previous studies have used different ML models (with LSTM being the most popular one) and various model performance metrics, making direct comparisons difficult. However, in most prior works, the wind speed was identified as the most important feature for good model performance but the effects of the other features have not been well studied. The power output predicted from the different models in Supplementary Table 1 is usually assessed by various model performance metrics based on their test data. This test data is an unseen portion by the model but consists of the same dataset that is used for model training and validation. Therefore, in most cases, the models are not deployed on other independent datasets, including datasets recorded at diverse future times or from other similar WTs. This limits a full appraisal of the methods for real-world deployment.  A recent survey by Yang *et al.* (2024) confirms the continuing shift towards deep and hybrid forecasting models while highlighting the importance of preprocessing, feature selection and computational cost.

From past studies, it has also been noted that the performance of different ML methods relies on numerous factors, including noise, sensor errors, data type, dataset size, dataset partitioning, data preprocessing, data structure, relationships between input and output features, learning method, algorithm selection and model optimisation. These factors can affect the robustness of the models, and their effects on the developed models need to be studied thoroughly. Data preprocessing through exploratory analysis is a determining factor for good model performance, although its importance has not always been highlighted in past research. Overall, the performance of a model is relative and must be assessed through a baseline model for benchmarking purposes before deploying the models in the real world to determine its performance, however, this also has not been a common practice in past research. Moreover, the input features commonly comprise multiple intrinsic parameters of the WT and SCADA data consisting of these WT parameters are often not readily available for each WT when deploying the model on another dataset. There is also a necessity to identify a suitable trade-off between the amount of data used, computational time required, model accuracy, and complexity for future applications as further supported by Ahmadi and Khashei,(2021). Recent transfer-learning research has begun to address cross-turbine deployment by selecting relevant source turbines and adapting a pretrained deep model to target turbines with limited data (Dong and Xiao, 2024).

Therefore, to address the above limitations, this study aims to explore multiple critical areas involved in the application of ML methods for power predictions in WT. The effect of the data volume, data preprocessing, feature types and selection and the use of a baseline model are investigated regarding the performance of an ANN model as applied to an existing real-life WT. To enhance the usefulness of the ML model developed for the end user, the energy outputs of the WT in 4-h time blocks for up to 72 hours in the future are calculated and visualised. The energy outputs are then used to identify optimal times to perform maintenance (which would be ideally scheduled during periods of low energy outputs). Furthermore, the energy saved when the maintenance is scheduled during low energy output periods instead of when it was actually carried out, is demonstrated. Finally, this research demonstrates how to use a separate weather dataset to broaden the real-world potential of the method while incorporating adequate data preprocessing, feature selection, data volume, and algorithm complexity for enhanced ML applications in the wind energy sector.

# 2 Methodology

This methodology section provides an overview of the different steps involved in ML model development and assessment. First, the principles of WT power generation are presented before an overview of the two datasets (dataset 1 from the WT and dataset 2 weather forecasts) used in this study. To gain insights into the dataset characteristics, exploratory data analysis and visualisation are conducted. For dataset 1, extensive preprocessing procedures were performed, including filtering based on the time taken for power to reach a steady state after startups and shutdowns, removal of brake states and power values less than zero, feature selection to identify the most significant features, resampling the data to a 4-min interval, outlier detection and removal to eliminate abnormal data points and data normalisation. Dataset 2 undergoes its own preprocessing, including feature relevance assessment, data alignment, and normalisation. The development of an LR baseline model using dataset 1 is carried out to serve as a benchmark against a more complex ANN model. Lastly, dataset 2 is used to deploy the ANN model, enabling WT power generation prediction and optimal maintenance scheduling (from an energy output perspective).

## 2.1 Theory/Calculation

To provide some initial useful insights into the power generated by WTs, equation (1) demonstrates the relationship between environmental conditions and power output. In theory, the amount of power generated by a WT is modelled as a function of the area swept by the rotor blades, air density, and the kinetic energy flux into the turbine rotor (Abolude and Zhou, 2018):

$$P = \frac{1}{2}\, rAU^3 C_p$$

where A is the air density, r is the rotor swept area, U is wind speed and $C_p$ is the coefficient of performance.

When considering a specific WT, the rotor swept area would be constant and the air density would depend on the air temperature, pressure, and humidity. The WT power output is also affected by wind speed, and this relationship is normally depicted by power curves for WTs. These can be used to estimate the wind energy potential at a candidate site (Lydia *et al.*, 2014), for performance monitoring, and even for outlier detection.

## 2.2 Overview of the Datasets

Two datasets were used in this study. Table 1 represents the parameters of the Levenmouth WT in Fife, Scotland (see also in Appendix Supplementary Figure 1) from which the SCADA and Met Mast data forming dataset 1 was obtained.

Table 1: Levenmouth Wind Turbine Characteristics (McKeever, 2020).

| Parameters of the WT | Values |
|---|---|
| Rotor diameter | 171.2 m |
| Capacity | 7 MW |
| Hub height | 110.6 m |
| Blade length | 83.5 m |
| Rated frequency | 50 Hz |
| Rotor speed | 5.9-10.6 rpm |
| Wind speed | 3.5-25 m/s |
| Rated wind speed | 10.9 m/s |

WT = Wind Turbine.

The METEO weather forecasts forming dataset 2 were obtained from the Intelligent Plant Appstore and a meteorological (MET) station located close (~14.5 miles) to the WT studied. This dataset was used as an independent set of data to predict the power from the models developed, for 72-h into the future and is considered as the deployment dataset in this study. The METEO forecasts were available as multiple structured data tables at 4-h intervals for different consecutive hourly timestamps up to 72 hours into the future. Once these tables were combined, one single table at a 1-h interval was obtained for weather forecasts for each hour in the future for up to 72-h into the future. The accuracy of weather forecasts in dataset 2 can be presented as a five-day forecast that can accurately predict the weather approximately 90% of the time (MET Office, 2022). Therefore, the accuracy of the power predictions from those weather forecasts may already include a small error, which can, however, be considered negligible (Shonk, 2018) .

Figure 1 represents the location of the datasets used in this study. Dataset 1 was from the WT located offshore in Levenmouth, Scotland and was purchased from Offshore Renewable Energy Catapult Ltd. Dataset 2 was obtained from the MET station as shown in Figure 1 (with the specific location on the map additionally provided in Appendix Supplementary Figure 2).

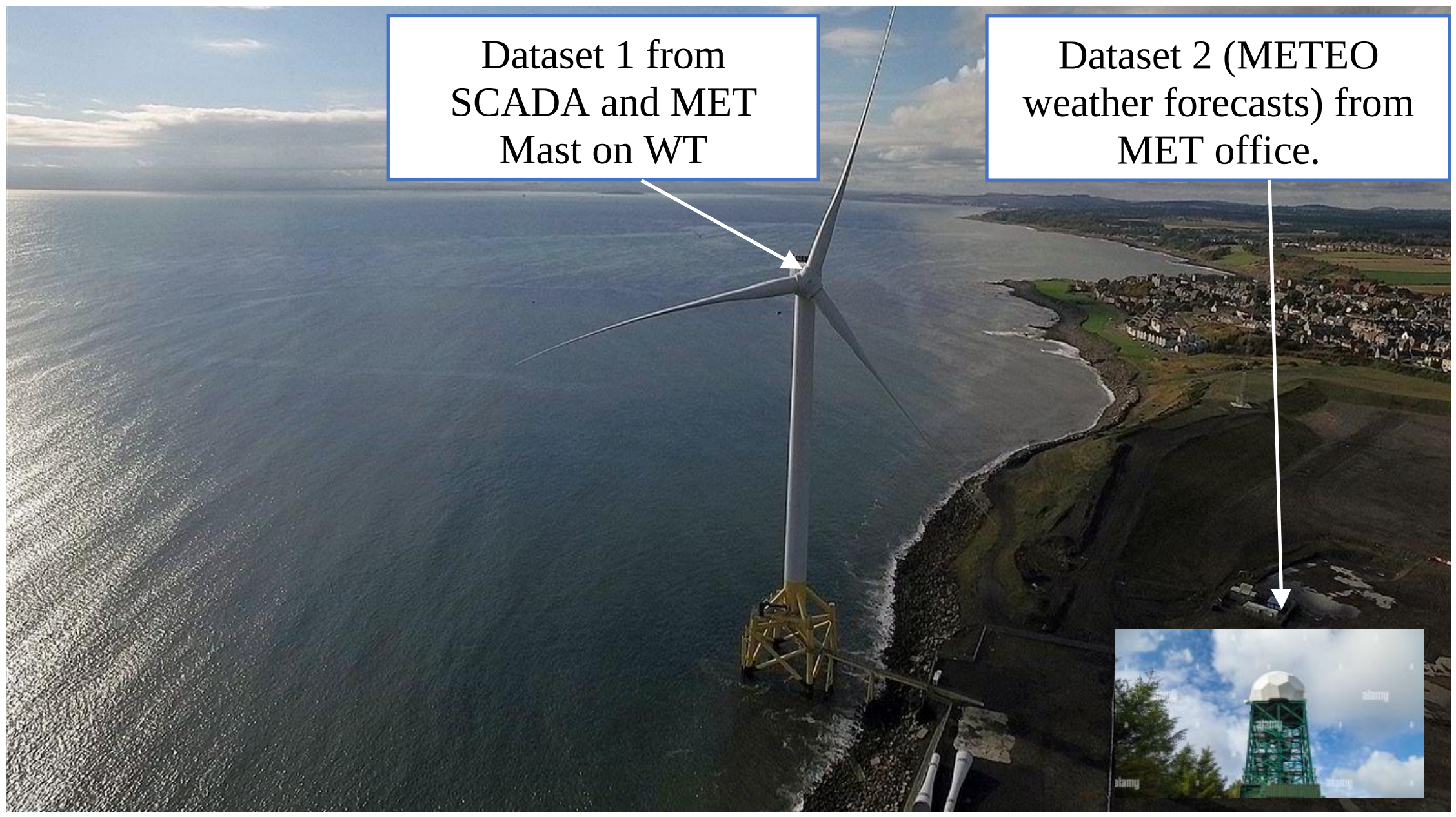

Fig. 1. Levenmouth Wind Turbine for SCADA, MET MAST (Snieckus and London, n.d) and MET office; Credit: Alamy Stock Photo, n.d).

Table 2 shows the characteristics of dataset 1 and 2 as obtained. Dataset 1 is used to train and test the models. Dataset 2 is used to deploy the models and assess their capability to perform in another scenario independent from the dataset used to train and test the model.

Table 2. Overview of datasets used in this study.

| **Dataset number** | **Dataset 1: SCADA and MET Mast** | **Dataset 2: Meteorological (METEO) Weather Forecasts** |
|---|---|---|
| **Components of dataset** | SCADA and MET MAST data on WT. | METEO dataset: Weather forecasts from a nearby MET office. |
| **Description** | WT sensors fitted on the WT to measure intrinsic and weather-related parameters recorded at different heights. | NWP for each hour in the future derived from multiple DataFrames available at 4-h intervals; up to 72-h in the future accessible |
| **Example of features present** | Pitch angle, Brake states, Nacelle orientation, Wind Speed, Wind direction, Atm pressure, Air temperature | Wind speed, Wind direction, Atm pressure, air temperature plus001hr, plus 002hrs etc. for 72-h in the future |

| **Number of features** | >100 | >5 |
|---|---|---|
| **Time period available** | 01-2019 to 12-2019 | Up to 60 days in the past and 72-h in the future |
| **Interval** | 1 sec | 1-h interval |
| **Number of datapoints (NDPs[2])** | ~30 million | 24823 |

WT = Wind Turbine, NDPs = Number of Datapoints, NWP = Numerical Weather Predictions, Atm = atmospheric.

## 2.3 Exploratory Data Analysis

A high-density scatter plot consists of a large concentration of data points, closely packed together or overlapping each other. These plots are useful when datasets consist of a large amount of data points and visualisation is required to identify patterns or trends. Figure 2 displays the power curve of dataset 1 as a high-density scatter plot and two regions of densities are observed. The first represents the main curve and has a pixel density of approximately 500 data points per pixel. The second category includes areas with a pixel density above 2500 data points per pixel and is present towards the bottom and top of the curve.

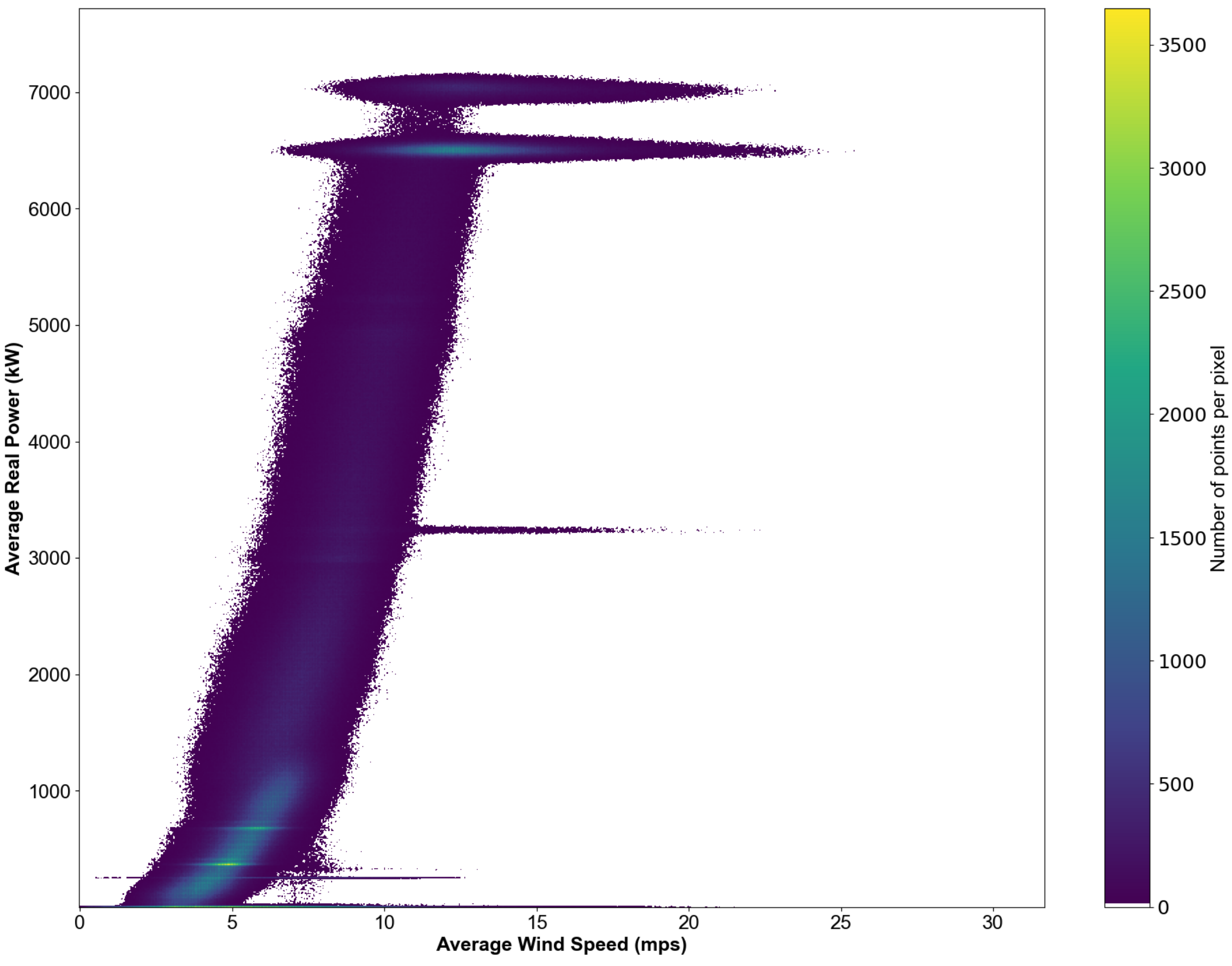

Fig. 2. High Density Scatter plot of power curve.

In the literature, it is common to distinguish between different types of outliers occurring on a power curve (Zou and Djokic, 2020) (Zhao *et al.*, 2018). Figure 3 displays the power curve with the four different types of outliers.

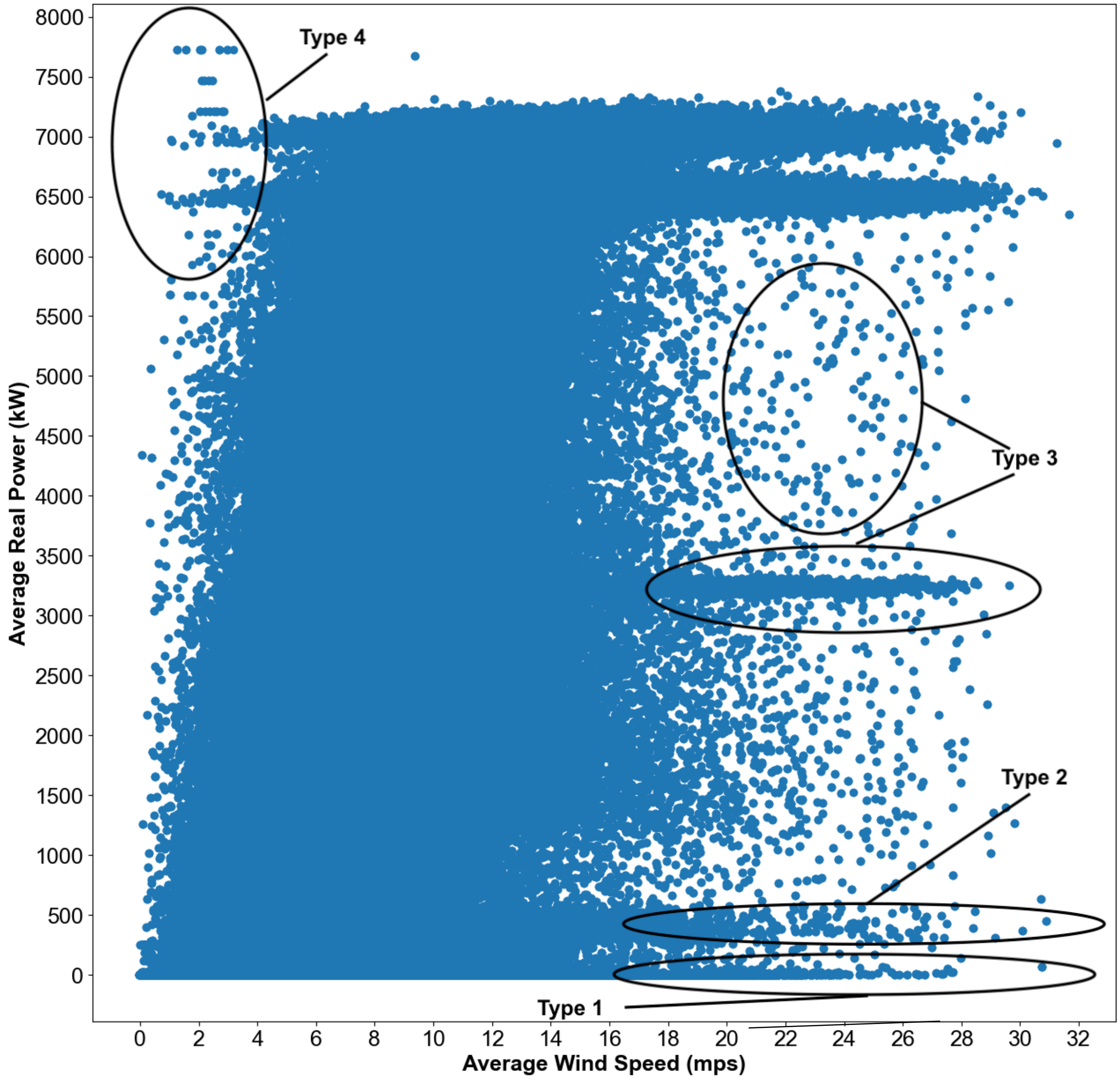

Fig. 3. Power curve of the Wind Turbine, with four different types of outliers highlighted.

Type 1 are bottom-stacked outliers. These are produced by failure or outage, unplanned WT maintenance, and faults in WT measurement or communication errors. Power output data that are close to or lower than zero are caused by wrong measurements or WT malfunctions. When the WT does not rotate, the power output is zero, however, if the WT's control system is still energised, the recorded power output might be negative (Zou and Djokic, 2020). Type 2 outliers in Figure 3 are due to wind curtailment, which is a set of control measures that limit the power output of WTs so that they are lower than normal levels (Xiaojun et al., 2019). Even if the wind speed exceeds the rated speed, the output power is maintained continuously at a certain point. Type 3 outliers are sparse outliers that typically occur because of extreme weather conditions, random noise, or a transition period when the WT is going from shutdown to startup. Type 4 outliers are top-curve stacked outliers arising from wind speed sensor failures. The visualisation of the power curve can be used to identify and eliminate outliers. In this study, outlier removal is performed according to the identification of the power curve outliers described above.

## 2.4 Preprocessing of Datasets 1 and 2

Figure 4 provides an overview of the different data preprocessing steps used in this work with the number of data points left after each preprocessing step. Details of each step are provided in subsequent subsections.

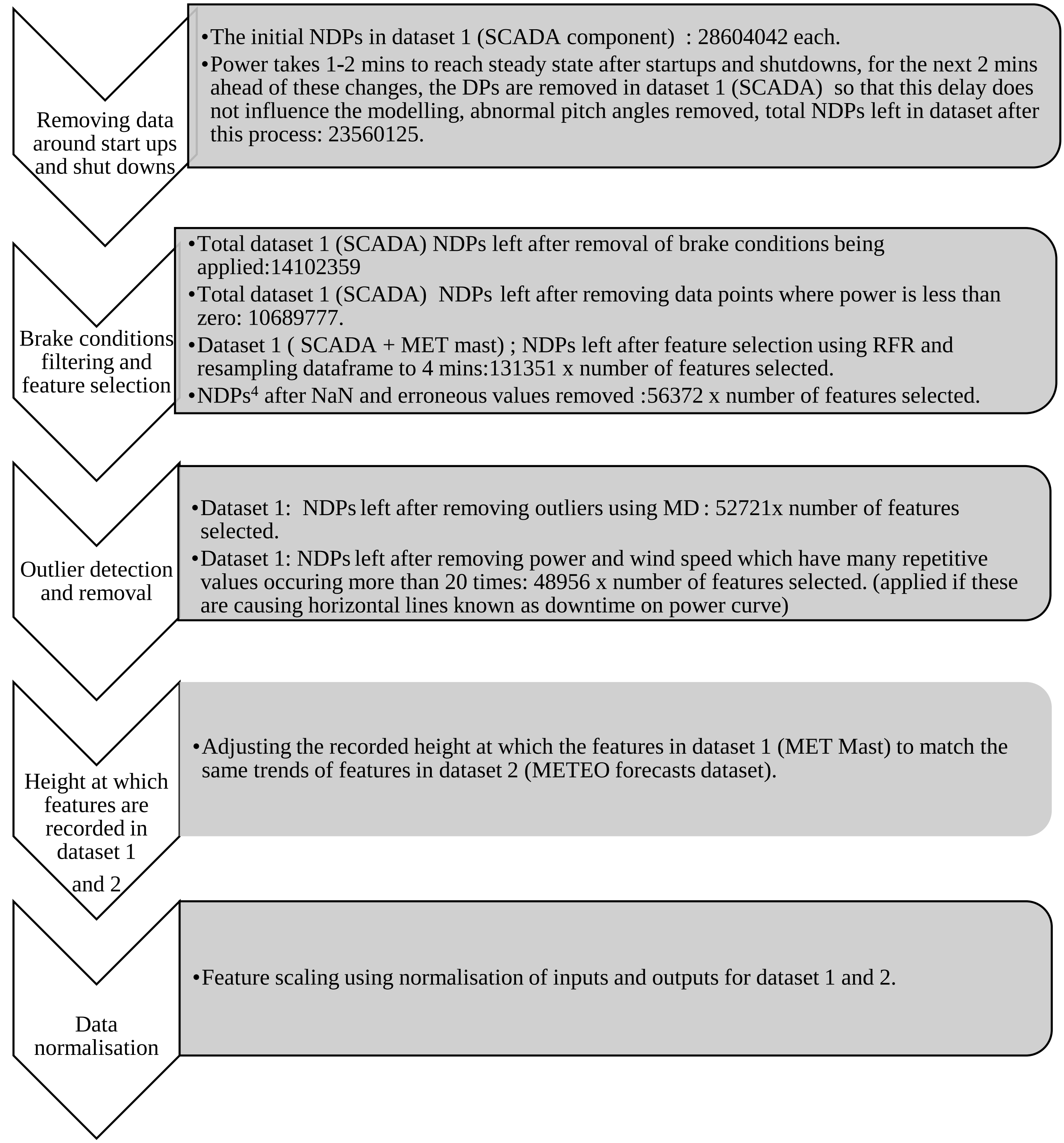

Fig. 4. Overview of preprocessing techniques used in this study: DP = Datapoints, NDPs = Number of Data points, WT = Wind Turbine, Dataset 1 = SCADA + Met Mast datasets on WT, Dataset 2 = METEO Weather Forecasts from MET station, MET Mast = environmental sensors on WT, mins =minutes, RFR = Random Forest Regressor, MD = Mahalanobis distance.

•

### 2.4.1 Data Filtering for Dataset 1

Visualisation of dataset 1 (SCADA component of the dataset) identified a change in power trends 2 mins after startups and shutdowns. This is associated with the time required to achieve steady conditions. Therefore, the data points for 2 mins after the startups and shutdowns have been removed. Moreover, if data points occurred at an abnormal pitch angle, they were also eliminated.

A further breakdown of dataset 1 (SCADA component of the dataset) shows that different brake features were being applied to influence the power output, with the most common brake states resulting in power equaling zero. The brake states were being applied for different possible reasons, including reducing the power output before a shutdown, wind curtailment, pitch control, active train damping, active load control and blade load monitoring. Values of 1 and 0 were assigned respectively to represent whether there was a brake being applied or not in dataset 1 (SCADA component of the dataset), and the resulting power output range was investigated for each of them. The number of data points consisting of power being less than or equal to zero was found to diminish consequently when all different brake features being applied were removed.
Power Output ($P_{out}$) resulting from setting the following brake feature conditions to none is shown in Table 3.

Table 3. Power output and data point removal due to brake conditions.

| Dataset 1 for 2019 Power output range (kW) | Initial NDPs in the specified power output range | Final NDPs in the power output range after the removal of data points with applied brake conditions | % of NDPs that have been removed due to applied brake states in the specific power output range. |
|---|---|---|---|
| **$P_{out}$ <0** | 2,890,572 | 1590 | 99 % |
| **$P_{out}$ > 0** | 11,211,765 | 10,688,187 | 4.6 % |

$P_{out}$ = Power output of Wind Turbine, NDPs = Number of Datapoints.

From Table 3, 99% of the data points of power output being less than zero in the dataset are occurring due to some brake conditions being applied. The 1% of power being less than zero may be occurring due to sensor errors.Therefore, all data points of power output being less than zero are removed from the dataset so that power predictions are not affected by the brake states.

### 2.4.2 Feature Selection for Dataset 1

Feature importance methods allocate a score to input features depending on their impact on predicting a target variable. The scores are used for a better understanding of the dataset and to reduce the number of input features, in addition to assisting with the interpretability of trained models. Rather than directly using the raw data points, extracting relevant features from the high-dimensional data can help reduce input dimensionality and focus on essential information for modelling. From an initial 900 features in dataset 1 (SCADA and Met Mast data), a Random Forest Regressor (RFR) was used to determine the number of features that have the largest impact on WT power output.

RFR uses an ensemble method that combines the quality of filter and wrapper methods, thus making them highly accurate and interpretable (Dubey, 2018). RFR measures feature importance by evaluating the decrease in the model's performance when a particular feature is randomly shuffled or removed from the dataset during the training process. The greater the drop in performance, the more important the feature is. The algorithm aggregates these individual feature importance scores from all decision trees in the random forest ensemble to provide an overall measure of feature importance. The resulting weather-related features that had the highest importance scores, used as inputs to the model were, wind speed, wind direction, air pressure, and air temperature.

### 2.4.3 Data Downsampling for Dataset 1

Excessive high-resolution data strains computer memory resources, particularly for complex models like ANNs. This hinders effective training due to increased computational demands and potential overfitting issues (Cho *et al.*, 2015). To address this, the data were aggregated from 1-sec granularity to 4-mins intervals, striking a balance between computational efficiency and model performance. This was performed after the initial steps of gap elimination, and feature retention. The dataset was resampled to a resolution of 4 mins intervals when applying MD to reduce the computational load of the outlier detection technique used while maintaining and capturing all the variability of the features’ values happening over time, which could affect model performance.

### 2.4.4 Outlier Detection for Dataset 1

The Mahalanobis distance (MD) is a measure of the distance between a point and a distribution. This method considers the covariance structure of the data, making it useful in scenarios where the features are correlated. MD can be calculated as the distance d from a vector y to a distribution with mean μ and covariance Σ (MathWorks, 2023) (Leys *et al.*, 2018) and is denoted in equation (2):

$$d=\sqrt{(y\text{-}\mu)\sum^{-1}(y\text{-}\mu)'} \qquad (2)$$

where d represents the distance of how far y is from the mean in the number of standard deviations, y is a vector of features $y=(y_1,y_2\ldots y_k)$, μ and Σ are the sample mean and covariance of the reference samples, respectively.

MD was used to remove any outliers from the power curve and can be observed in Figure 5, where the outliers are orange and the remaining data points are blue. Other methods such as Euclidean distance, box and plot, elliptical envelope, and Z-score were explored on a trial-and-error basis and MD was found to be the most effective in removing outliers. MD worked better than the z-score and Euclidean distance for outlier detection in this case because it considers the covariance structure of the data, handles unequal variable variances, and is robust to outliers (Cansiz, 2021).

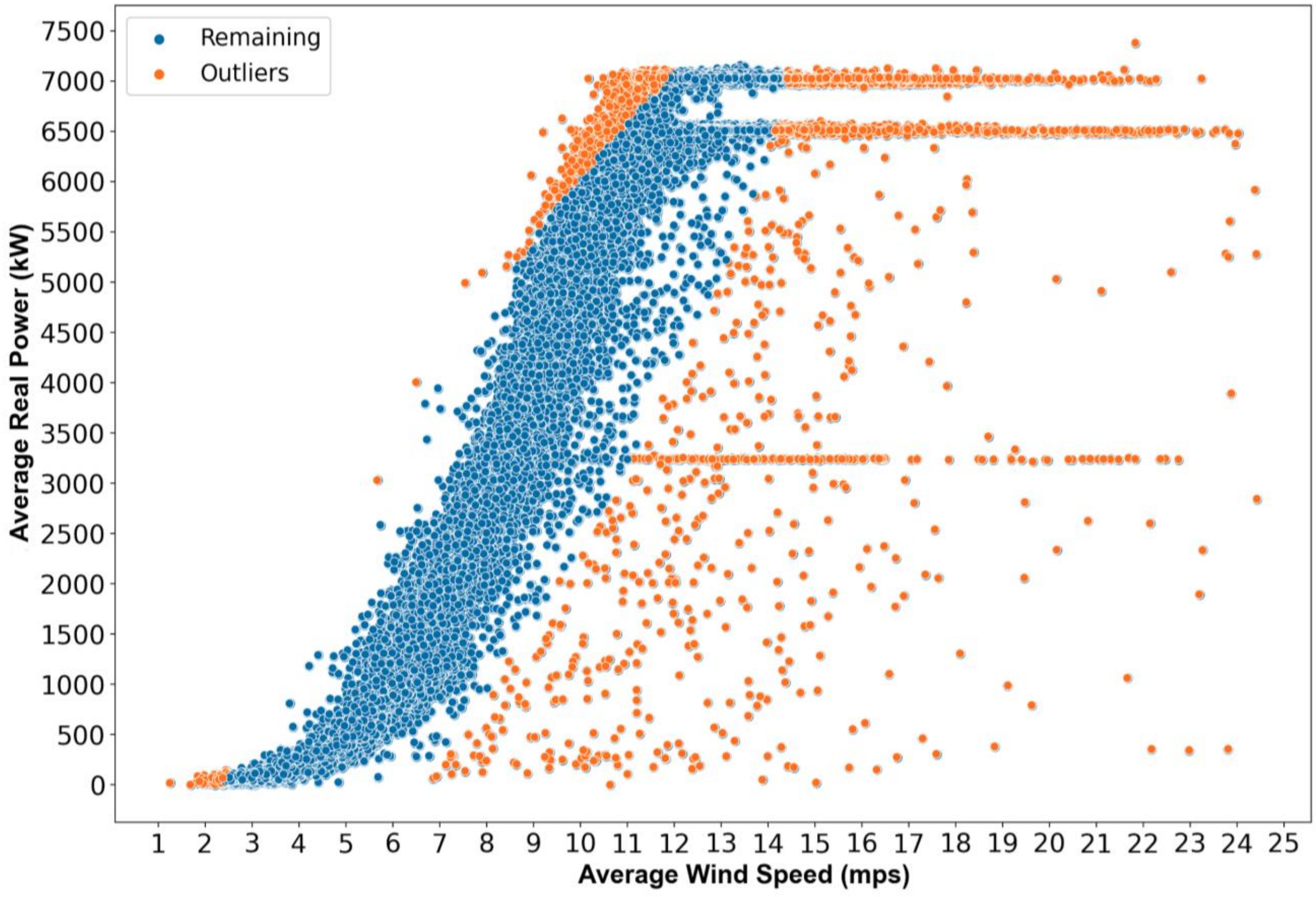

Fig. 5. Average real power (kW) against Average wind speed (mps) after applying Mahalanobis distance.

### 2.4.5 Data Normalisation for Dataset 1

Data normalisation ensures that all the model input features are on the same scale and helps to enhance the performance and reliability of a machine learning model by reducing the possibility of an individual feature having more importance than it should due to it having a larger data range. (Jo, 2019). Equation (3) was used for data normalisation in this work:

$$X_{norm}=\frac{X\text{-}X_{min}}{X_{max}\text{-}\ X_{min}} \tag{3}$$

where X is the value in the DataFrame, $X_{norm}$ is the normalised value, $X_{min}$ is the minimum value of all the values of a specific feature, $X_{max}$ is the maximum value of all the values of a specific feature.

#### 2.4.6 Preprocessing of Dataset 2

The preprocessing techniques applied to dataset 2 (METEO forecasts) are detailed as follows. To test the model with dataset 2 (METEO weather forecasts), the same features used to train and test the model using dataset 1 were obtained. These were wind speed, wind direction, air temperature and atmospheric pressure. Dataset 2 was then aligned with dataset 1. Key features (e.g., wind spend) of historical weather conditions in the SCADA data and Environmental MET Mast data (forming dataset 1) were compared with the METEO weather forecasts (dataset 2) for the same time period. When the same features from datasets 1 and 2 were plotted for the same time period, a constant difference in the values for similar features was observed because the weather data were being recorded at different heights for datasets 1 and 2. The affected weather values in dataset 2 were adjusted by using a constant value to account for this. Dataset 2 was normalised using the same method as that used for dataset 1.

### 2.5 Model Development

A baseline model using LR was first developed to predict WT power from weather data before a more complex ANN model. An overview of the model development, deployment inputs and outputs for both models is provided in Figure 6.

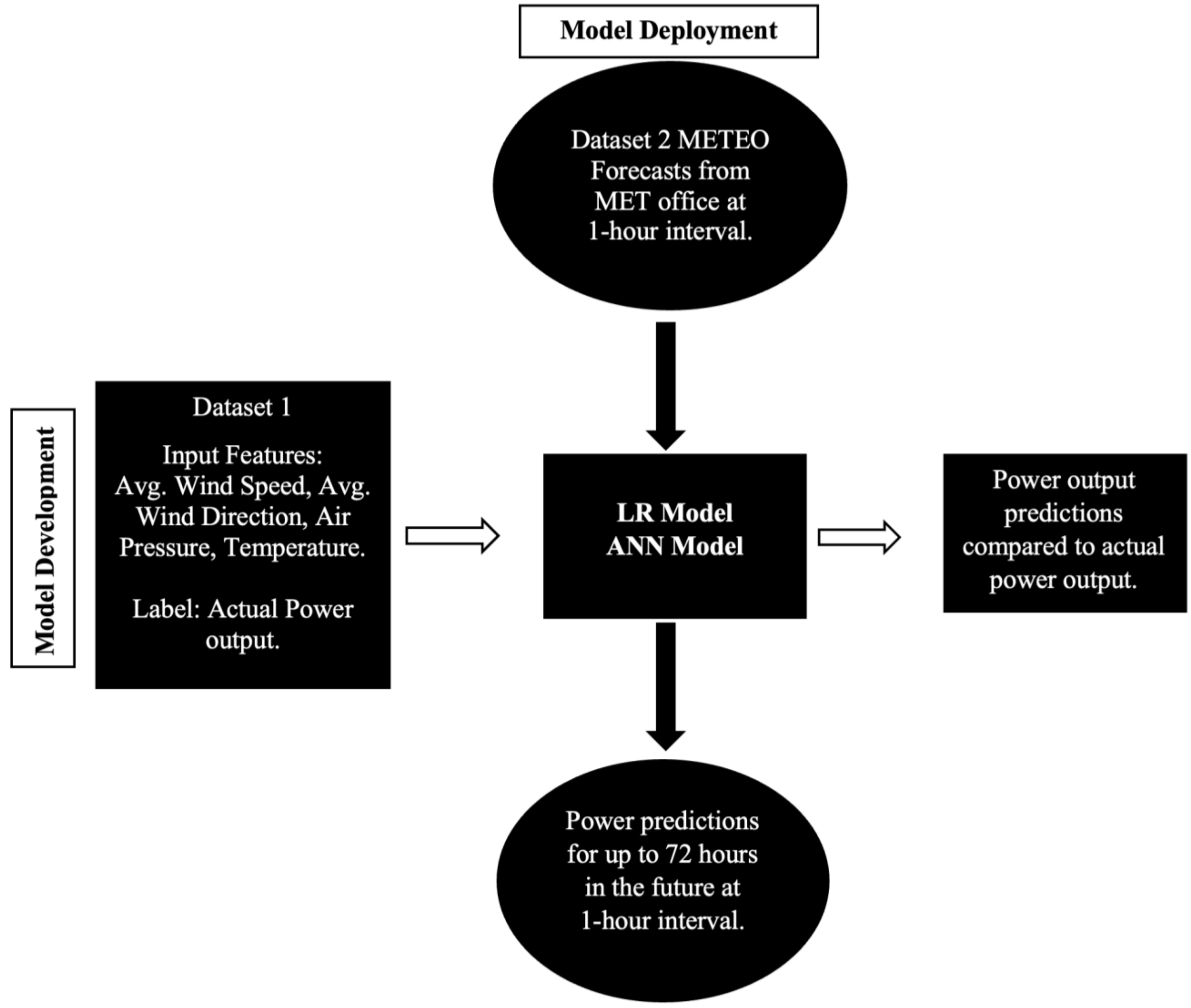

Fig. 6. Overview of model development and deployment: Avg.= Average, ANN = Artificial Neural Network, WT = Wind Turbine, LR = Linear Regression.

### 2.5.1 Baseline Model

A baseline LR model was developed using the scikit_learn library (Scikit learn, 2023). The goal of LR is to minimise the residual sum of squares between the targets observed in the dataset and the predicted targets. Only one input feature was used in the LR baseline model, and this was wind speed as the RFR feature selection process identified this as the most important feature.
The equation (4) for LR is provided below (Scikit learn, 2023) (P.Fabianpedregosa et al., 2011):

$$Y_i = \beta_o + \beta_1 X_i + \varepsilon_i \tag{4}$$

where $Y_i$ is the dependent feature, $\beta_o$ is the population Y-intercept, $\beta_1$ the population slope coefficient, $X_i$ is the independent feature and $\varepsilon_I$ is the random error term.

### 2.5.2 ANN Model

To choose an appropriate ML model for the datasets available, the volume of data is an important factor that should be taken into consideration. Since the dataset consists of approximately 50,000 data points and multiple features after preprocessing, an ANN with several hidden layers seemed appropriate, as they are able to perform well on large datasets and can generalise well on unseen data (Alwosheel, van Cranenburgh and Chorus, 2018).

A feedforward ANN was developed in TensorFlow (Mart´ın *et al.*, 2016) and Keras (Chollet and al., 2015). The ANN uses a back-propagation algorithm to update the weights for error minimisation. The output of a fitted ANN is given by (Ascher *et al.*, 2022) equation (5):

$$y_k = f_o\left(\sum_h W_{hk}\, f_h\left(\sum_i w_{ih} x_i\right)\right) \tag{5}$$

where $y_k$ is the network's estimation of the response feature. The products of the weights W for I inputs features x for the hidden nodes h are summed. These are transferred by the activation functions $f_o$ for the output and$f_h$ hidden nodes.

#### 2.5.2.1 ANN Optimisation and Performance Assessment

Dataset 1 was shuffled to remove seasonal patterns in the time series data and then partitioned into a 60:20:20 ratio for training, validation, and testing, as this ratio has been commonly used in previous studies (Gholamy, Kreinovich and Kosheleva, 2018). However, varying the train-test-split to 40:30:30 was also investigated and did not significantly affect the performance of the model. Training allows the model to learn from labelled data, validation enables fine-tuning of the model's hyperparameters and testing assesses the model's performance on unseen data (data not used in training or validation) to gauge its ability to make accurate predictions in practical scenarios. Model hyperparameters are components of a model that the developer selects and are not determined through training processes. Hyperparameter tuning is important for determining the optimal values to develop the highest-performing model. To perform hyperparameter tuning, a grid search was used. The KerasRegressor from the keras.wrappers.scikit_learn module was used to integrate the ANN model into the scikit-learn machine learning framework for a grid search within the Python programming language (P.Fabianpedregosa et al., 2011).

A grid search was performed with three and five-fold cross-validation. The hyperparameter search space was explored with different batch sizes, epochs, learning rates, the number of neurons per hidden layer, and the number of hidden layers themselves to find the best network configuration. From reviewing the literature, several combinations were assessed starting with a single hidden layer and up to five hidden layers. The number of neurons and learning rate were also varied until the optimum values were identified. The output layer transfer function was set as sigmoid as this keeps all the values between 0 and 1, thus preventing any negative values.

The model was first trained on one month of data from dataset 1 which was March 2019 and then on one year of data from dataset 1 which was from Jan 2019 to Dec 2019, to investigate the effect of data volume on model performance. The combination that provided the best-estimated performance after a grid search of hyperparameters and cross-validation is shown in Table 4.

Table 4. ANN Model hyperparameters.

| Hyperparameter | Optimised Hyperparameter Values |
|---|---|
| **Validation split** | 0.1 |
| **Number of hidden layers** | 2 |
| **Number of neurons in the first hidden layer** | 80 |
| **Number of neurons in the second hidden layer** | 60 |
| **Learning rate** | 0.002 |
| **Number of epochs** | 200 |
| **Batch size** | 10 |

| **Optimizer** | Adam |
|---|---|
| **Activation function** | Relu |
| **Output layer transfer function** | Sigmoid |
| **Dropout** | 0.01 |

The predictive performance of the models was assessed in terms of the coefficient of determination ($R^2$), Mean Absolute Error (MAE), Root Mean Squared Error (RMSE), and its normalised version (NRMSE), all of which are presented in equations 6-9. $R^2$ calculates the amount of variance in the predictions explained by the dataset. $R^2$ can range from 0 to 1, and the closer it is to 1, the better the goodness of fit.

$$R^2=1-\frac{\sum_{i=1}^{n}(Y_i-\hat{Y}_i)^2}{\sum_{i=1}^{n}(Y-_{i}Y_i)^2}$$

The MAE is the mean of the magnitude of the differences between predicted and actual values and is domain specific. Ideally, lower MAE values in the range of the actual values to be predicted are most desirable.

$$MAE=\frac{1}{n}\sum_{i=1}^{n}|Y_i-\hat{Y}_i|$$

RMSE and its normalised version NRMSE is the standard deviation of the prediction errors and ideally should be as low as possible.

$$RMSE=\sqrt{\frac{1}{n}\sum_{i=1}^{n}(Y_i-\hat{Y}_i)^2}$$

$$NRMSE=\frac{RMSE}{mean\ \hat{y}}$$

For all equations 6-9 $Y_i$: ground-truth value, $\hat{Y}_i$: predicted value from the model, and n: number of datums.

### 2.6 Model Deployment- Maintenance Scheduling

At the time of deployment of the model, the METEO forecasts from 01-01-2022 to 01-03-2022 forming dataset 2 were available (as these were the forecasts for 60 days) and were used to simulate the power predictions for the future when the weather forecasts are available. The predicted power outputs (using dataset 2 as inputs to the trained model) were compared to the actual power outputs of the WT from dataset 1 using MAE. A feature which keeps track of when maintenance crew were present at the WT was identified in dataset 1 (SCADA component). This was used to calculate the average time for a maintenance event which was 3 - 4 h. Therefore, the power predictions were used to determine the total power generation during 4-h periods and identify periods of low power generation, ideal for engineers to perform maintenance.

## 3 Results and Discussion

This section presents the results of the different models developed in this study. The model complexity, training data volume, input features and data preprocessing are explored. To provide practical insights for end-users, the average energy output in 4-hr time blocks was computed to identify future time periods of low power generation, that are more suitable for scheduling maintenance.

### 3.1 Power Predictions for 1 Month on Data: Baseline LR and ANN Models

Table 5 shows the model performance metrics calculated from the test set on the baseline LR and ANN models trained on one month of data (03-2019 consisting of 579 rows ×5 columns). Dataset 1 (SCADA and MET Mast) was used for training, validating, and testing. Dataset 2 (METEO forecasts) was used for the deployment of the model.

Table 5. Test set model performance metrics and model deployment error for different models using one month of data (03- 2019).

| **Input Features** | **Preprocessing Techniques** | **Model Used** | **Model Performance Metrics** | **Error Between the Actual Power from Dataset 1 (SCADA) and Predicted Power from Dataset 2 (METEO)** |
|---|---|---|---|---|
| Wind speed | -Data points are removed until power becomes steady after startups and shutdowns.<br>-Negative power removed.<br>-Brake conditions removed.<br>-METEO and MET trends attributes adjusted.<br>-MD for outlier removal. | LR<br>Baseline model | $R^2$: 0.81<br>MAE: 694.34<br>RMSE: 972.34<br>NMRSE:33.60 % | MAE>3000<br>RMSE: 3000<br>NRMSE: 43% |
| Wind speed, Wind direction, Air temperature, Atm. Pressure. | -Data points are removed until power becomes steady after startups and shutdowns.<br>-Negative power removed.<br>-Brake conditions removed.<br>-METEO and MET trends attributes adjusted<br>-MD for outlier removal. | ANN with Keras wrapper for three-fold cross-validation and hyperparameter tuning | $R^2$: 0.94<br>MAE: 264.28<br>RMSE: 524.91<br>NMRSE: 21.09% | MAE: 2161<br>RMSE: 2608<br>NRMSE: 37% |

ANN=Artificial Neural Network, Atm=Atmospheric, MD=Mahalanobis distance, RMSE=Root Mean Square Error, LR=Linear Regression, NMRSE=Normalised Root Mean Square Error, MAE=Mean Absolute Error, $R^2$=Coefficient of determination.

From Table 5, the ANN model performed better than the baseline model because the $R^2$ value was closer to 1 and the error was lower. ANNs have numerous advantages such as strong learning ability, increased functionality, and accuracy compared with simple LR models. However, the NRMSE between the actual power output from the SCADA data and the predicted power from dataset 2 (METEO weather forecasts) is only 6% different between the two models, questioning the value of using a more complex model. The MAE also shows a clear indication of the difference in the performance of the models when preprocessing steps are applied along with the RMSE and NRMSE.

### 3.2 Power Prediction for One Year of Data: Baseline LR and ANN Models.

Table 6 displays the model performance metrics for the entire dataset available (one year of data, 01-01-2019 to 31-12-2019 consisting of 48956 rows×5 columns). Dataset 1 (SCADA and MET Mast) was used for training, validation, and testing. Dataset 2 (METEO Forecasts) was used for deployment.

Table 6.Test set model performance metrics results and model deployment error for LR models and ANN models using one year of data.

| **Input Features** | **Preprocessing Techniques** | **Model Used** | **Model Performance Metrics** | **Error Between the Actual Power from Dataset 1 (SCADA) and Predicted Power from Dataset 2 (METEO Forecasts)** |
|---|---|---|---|---|
| Wind speed | None | LR Baseline | $R^2$: 0.31 MAE:1202.9 RMSE:1663 NRMSE:128% | MAE between power from dataset 1 (SCADA) and power predicted from dataset 2 (METEO forecasts): >3000 |
| Wind speed | -Data points are removed until power becomes steady after startups and shutdowns. -Negative power removed. -Brake conditions removed. -METEO and MET Mast trends attributes adjusted. -MD for outlier removal. -Excessively repetitive values removed. | LR Baseline | $R^2$: 0.94 MAE: 441.1 RMSE: 544.7 NRMSE: 22% | MAE between power from dataset 1 (SCADA) and power predicted from dataset 2 (METEO forecasts): 2072.99 |
| Wind speed, Wind direction, Air temperature, Atm pressure. | None | ANN with Keras wrapper for 5-fold cross-validation and hyperparameter tuning | $R^2$: 0.74 MAE: 596 RMSE: 1052 NRMSE: 83% | MAE between power from dataset 1 (SCADA) and power predicted from dataset 2 (METEO forecasts): > 3000 |
| Wind speed, Wind direction, Temperature, | -Discarding data points 2 mins after shutdowns and startups for power to reach steady state. -Negative power removed. -Brake conditions removed. | ANN with Keras wrapper for three-fold cross-validation and hyperparameter tuning | $R^2$: 0.98 RMSE: 283 MAE: 194 NRMSE: 11% | MAE between power from dataset 1 (SCADA) and power predicted from dataset 2 (METEO forecasts): 1462.17 |
| Atm Pressure. | -MD outlier removal.<br>-Repetitive values and NaNs removed.<br>-Matching trends of METEO forecast data and SCADA/MET datasets. | | | |

ANN=Artificial Neural Network, NDPs=Number of Data points, MD=Mahalanobis Distance, $R^2$=Coefficient of Determination, MAE=Mean Absolute Error, RMSE=Root Mean Square Error, NRMSE=Normalised Root Mean Square Error.

The NRMSE was lower in Table 6 when one year of data was used for the LR (22%) and ANN (11%) models (with data preprocessing), compared with the model trained on one month of data (Table 5), where the NRMSE was 33.6 % for the LR model and 21.09 % for the ANN model. This is because a larger volume of data captures more variability in real-world conditions, enabling better predictions. Features other than wind speed have played a role in developing models with better performance, although wind speed remains the most important and influential. Additionally, it was found that, under the same preprocessing techniques for both the ANN and the baseline model, the test set MAE for the ANN is 194, which is lower than the MAE for the regression baseline model, amounting to 441 (Table 6). The MAE when comparing the actual power from the SCADA and predicted power output from the METEO is also lower for the ANN. This shows that the ANN is a better model for power predictions from other independent datasets.

From Table 6, the ANN model was compared to the baseline LR model with a different number of inputs and preprocessing steps. Although more inputs improved the performance of the model, the LR model with one input (wind speed) was still found to perform relatively well. Table 6 also highlights the importance of effective data preprocessing for the ANN model, where models with preprocessing achieved higher accuracies and lower errors between actual and predicted power from different datasets. It has to be noted that the preprocessing techniques have been kept similar when training the ANN model with one month and one year of data, respectively. The same modelling hyperparameters were also applied, however, the number of folds for the cross validation was reduced from 5 to 3 when increasing the data from one month to one year since the dataset was larger and already showed good performance with three-fold cross validation. This reduction saved significant computational time.

Figure 7 compares model predictions (using dataset 2 as inputs) with actual results from the WT for a month period in 2022. Missing data points for the actual power are a result of the brake being applied (see section 2.4). Figure 7 shows that the trends of the predicted and actual power outputs are similar, and the average power difference was approximately 600 kW.

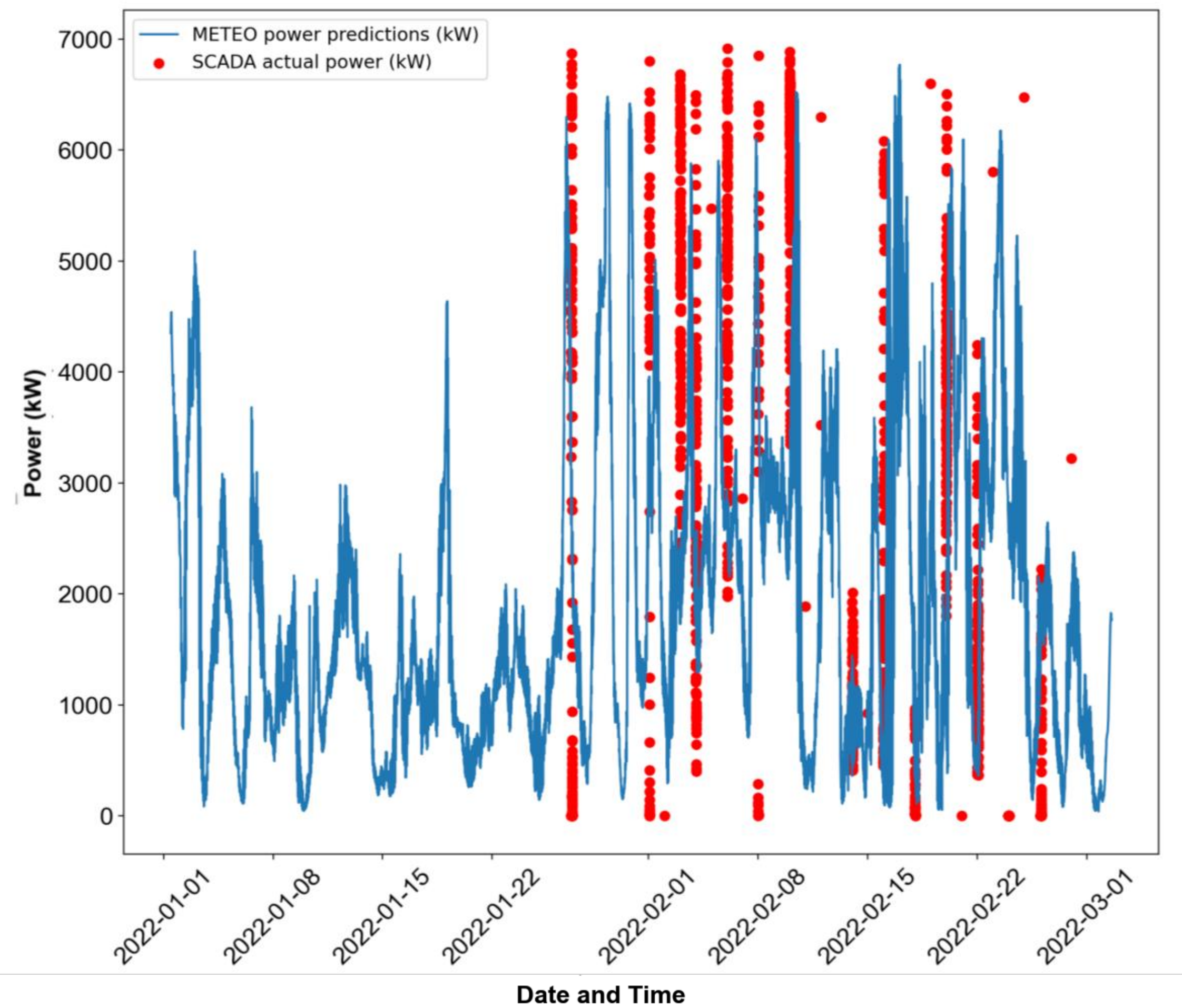

Fig. 7. Comparison of actual power output with power predictions from METEO dataset with format Year-Month-Day.

It is difficult to directly compare the results in this study to past research (Supplementary Table 1) because different metrics to assess model performance were often used. The MAE, for example, depends on the range of the target variable. For this study, the percentage error was 6.7%, which is calculated by dividing the MAE (194) over the range of the target variable (7000 kW – nominal power output) multiplied by a capacity factor (40%). Rashid et al., 2020 (Supplementary Table 1) achieved a percentage error of 3.6% for an MAE of 30 over a power output range of 2050 kW, for the same capacity factor. Khan et al., (2019) and Janssens et al., (2016) did not specify the actual range of values in their study so cannot be easily compared to other studies. RMSE also does not perform well if comparing a model's fit for different response variables. The NRMSE, however, can overcome these issues and can be used to compare the results in this work to others. In the current work, the NRMSE is 11% (ANN model with preprocessing), compared to Li et al., (2016) where it was approximately 21 % and Meka et al., (2021), where it was 19.7 %.

The $R^2$ can also be compared to determine how well the training data fits the model. In this study, the $R^2$ value of the ANN model with preprocessing was 0.98, which is very close to 1. This is comparable to Bilal et al., (2018) (0.986) and even outperforms others that have comparable amounts of data and inputs, such as Vladislavleva et al., (2013) whose $R_2$ was 0.85, Lin & Liu, (2020) 0.90, and Singh et al., (2021). If the $R^2$ in the current work is compared with that of Clifton et al., (2013), who used a synthetic dataset, it was 1.5 % lower. The authors believe that the model in this work performed better because the study takes into consideration a large amount of data, different preprocessing techniques and a well optimised model, in addition to having the capability to utilise other weather datasets in its deployment.

## 3.3 Power Predictions for 4-h Blocks of Time

Figure 8 shows the power against time in MM-DD HH (Month-Day:Hour) format for a time period in 2019. For the time period 10-07-05 (M-D-H) to 10-07-09 the average energy output was approximately 5126 kW. As this energy output, on average, was similar for several hours preceding 05:00, approximately 5000 kW was lost when the maintenance team was at the WT for the shutdown from 9:00 to 13:00 (red data points at 1 (second Y axis) Figure 8). If the maintenance team used the predictive model developed in this work and performed maintenance on the same day from 13:00 to 17:00, they would save approximately 2000 kW of power as the average energy output for that time period was approximately 3000 kW.

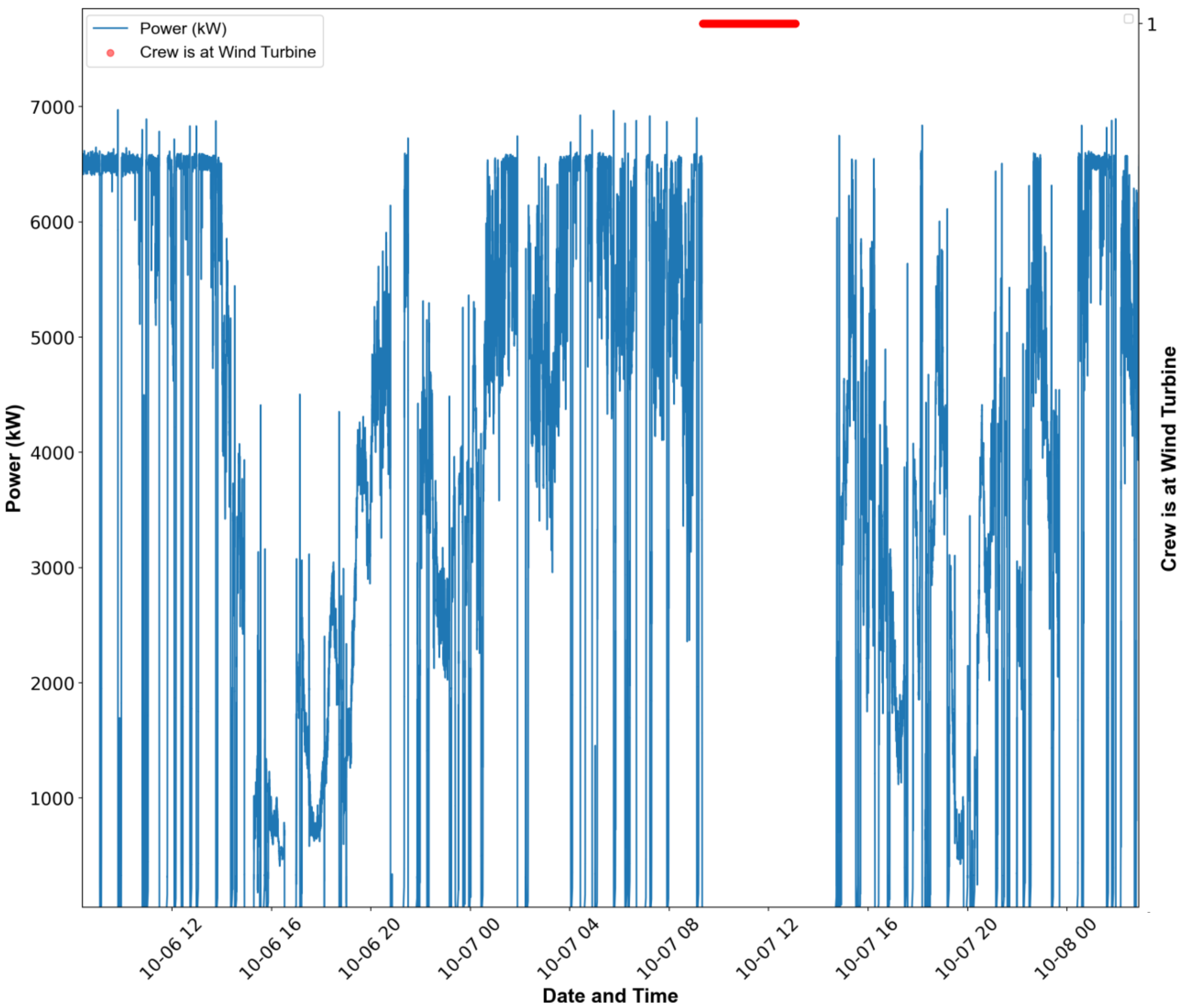

Fig. 8. Power output and maintenance crew at Wind Turbine for a 30-h time period in October with format Month-Day Hour; Key 0 = Crew not present, 1 = Crew present.

Figure 9 displays the predicted average energy produced from the WT over a 4-h period. The results are from the ANN model with the appropriate preprocessing techniques trained on one year of data. The numerous 4-h bars overlap each other without any gaps present during the timespan to provide a continuous overview, thus enabling the identification of time periods with low power output, where maintenance should be scheduled.

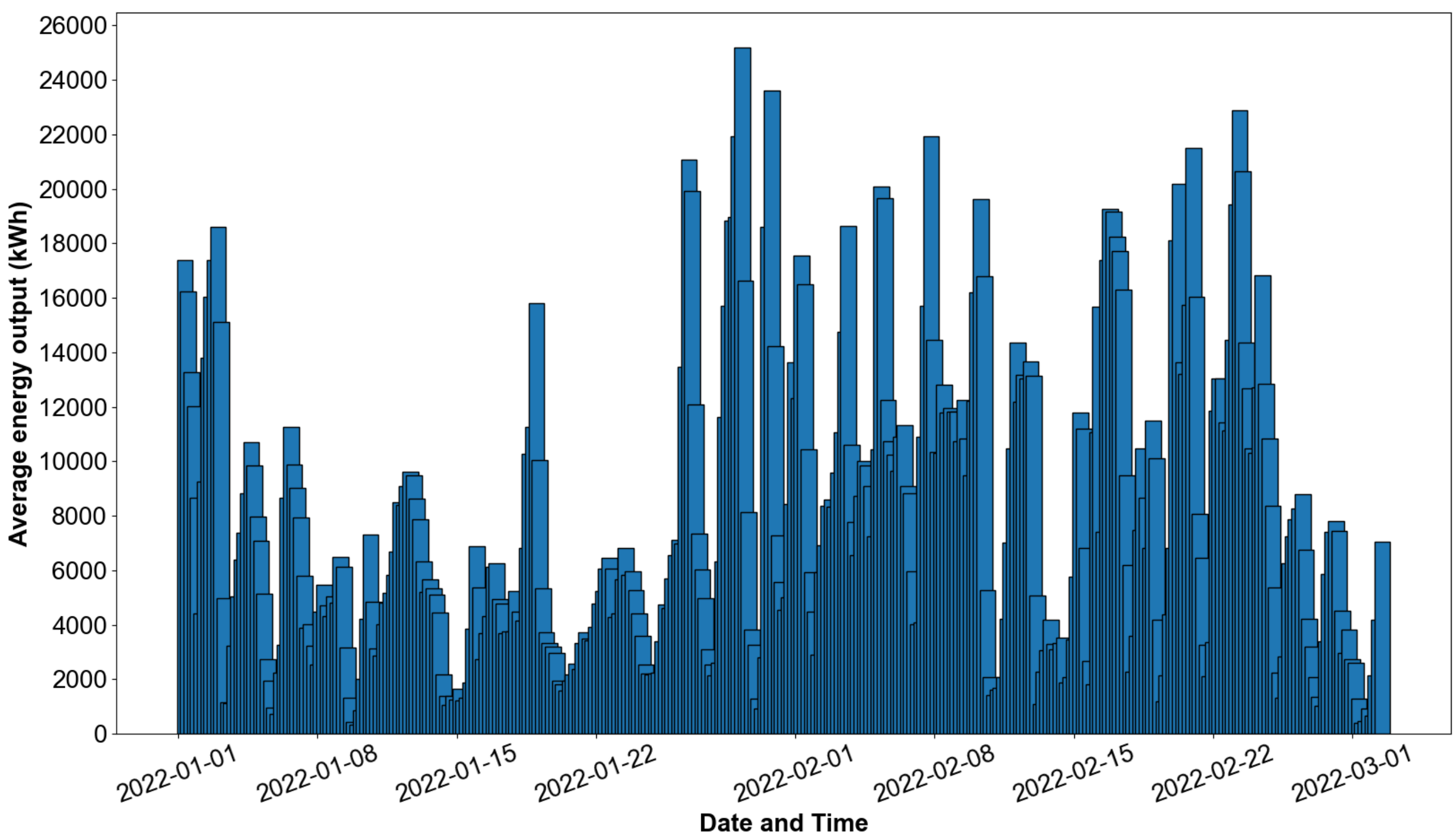

Fig. 9. Predicted average energy output in 4-h blocks over a period of 60 days with format Year-Month-Day.

The maximum average energy output for a 4-h block was over 24000 KWh and the minimum was approximately 500 kWh (Figure 9). In this period of three months, there were approximately eight periods of low (e.g., below 1000 kWh) average energy output that were distributed fairly evenly. If the maintenance engineers are informed about these times in advance, they have enough time to plan and assign tasks to their team members.

Figure 10 has the same results as Figure 9 but over a shorter (72 h) time period. Since the results in Figure 9 represent the predicted average energy output using weather forecasts for the past 60 days, at the time the model was deployed, it can be used to zoom in to 72 h from a date that is considered as the present to determine the optimal time to schedule maintenance in the near future. Considering the date 2022-02-07, blocks of 4 h of energy output are represented during that time as MM-DD: HH in Figure 9. For example, the lowest average energy output during this time span would be on 03-01 (MM-DD) from 04 am to 08 am, as shown on the x-axis of Figure 10. Therefore, the maintenance engineers should schedule maintenance during that time period. This information empowers users to identify periods of low power predictions, facilitating informed decisions for conducting maintenance on WTs.

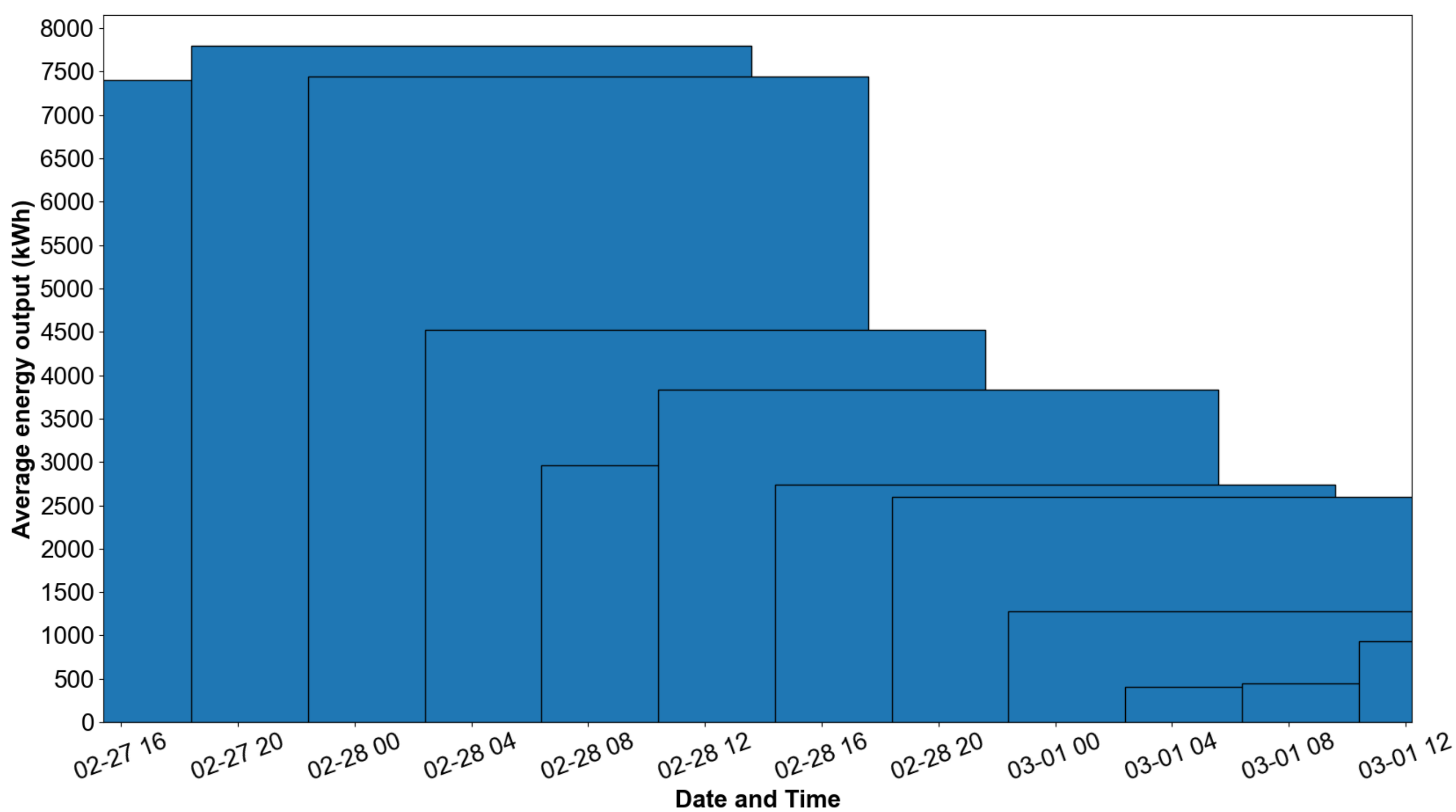

Fig. 10. Average energy output in 4-h blocks over a period of 72 hours with format Month-Day Hour.

# 4 Conclusion

In this study, an ML pipeline was designed to forecast WT power production for optimal maintenance scheduling. The datasets comprised of SCADA data collected over one month and one year on the WT, MET Mast weather conditions at the WT height (both forming dataset 1), and METEO weather forecasts (dataset 2) from a nearby weather station. Data preprocessing, such as outlier removal using the MD method, followed by feature selection using RFR and normalisation were performed. An LR baseline regression model was developed for benchmarking, and an ANN model was trained in TensorFlow with hyperparameter tuning using the Keras API. This fine-tuned configuration considerably improved the model accuracy, as demonstrated by the MAE, R2, and NRMSE metrics.

The effect of the preprocessing techniques for the baseline regression model indicated a large change in $R^2$ value from 0.31 to 0.94, demonstrating the importance of data preprocessing even on the simplest benchmarked model. The volume of data also impacted the performance of both the LR and ANN models. The baseline model improved from an $R^2$ of 0.81 to an $R^2$ value of 0.94 and an NMRSE of 33.60 % to an NRMSE of 22 %, when the volume of training data increased from a month to a year. The ANN improved from $R^2$ score of 0.94 to 0.98 and an NMRSE of 21.09 % to an NRMSE of 11 % for the larger volume of data. When comparing the baseline regression model with the ANN model for the same preprocessing techniques on one year of data, the ANN model had a lower NRMSE (11% compared to 22%) and deployment error which was 610.83 less than the baseline model.

The total energy output in a 4-h period that could be saved if maintenance was timed when the power output was low was shown to be approximately 2000 kW, in an example taken from dataset 1. Power predictions from the ANN model were deployed on weather forecasts to calculate average energy outputs in 4-h blocks and identify periods of low energy production. These predictions can help maintenance engineers at any similar offshore WT to plan and perform maintenance when the power output is lowest, thus minimising disruption to WT energy output.

# 5 Data Availability

Datasets related to this article can be found at [https://pod.ore.catapult.org.uk/data-collection/ldt-met-mast-scada-1sec], hosted at [ORE CATAPULT].

# 6 Declaration of Competing Interest

The authors declare that they have no known competing financial interests or personal relationships that could have appeared to influence the work reported in this paper.

# 7 Data Availability

Datasets related to this article can be found at [https://pod.ore.catapult.org.uk/data-collection/ldt-met-mast-scada-1sec], hosted at [ORE CATAPULT].

# 8 Declaration of Competing Interest

The authors declare that they have no known competing financial interests or personal relationships that could have appeared to influence the work reported in this paper.

# 9 Acknowledgements

This work was supported by UK Research and Innovation (UKRI) through EPSRC research grant number EP/S022996/1. The authors extend their appreciation to Intelligent Plant for also funding this work. I. Triguero holds a Maria Zambrano Fellowship at the University of Granada, and his work is also partly supported by the Spanish projects A-TIC-434-UGR20 and PID2020-119478GB-I00.